\documentclass{article}
\usepackage{spconf,amsmath,epsfig}
\usepackage{tabularx}
\usepackage{multicol}
\usepackage{multirow}
\usepackage{float}
\usepackage{amssymb}
\usepackage{mathrsfs}
\usepackage{xfrac}
\usepackage{array}

\begin{document}\sloppy

\def\x{{\mathbf x}}
\def\L{{\cal L}}

\title{Real-time Indoor Scene Reconstruction with RGBD and Inertia Input}
%
\twoauthors
  {Zunjie~Zhu}
	{Hangzhou Dianzi University}
  {Feng Xu}
	{Tsinghua University}



\maketitle

\begin{abstract}
Camera motion estimation is a key technique for 3D scene reconstruction and Simultaneous localization and mapping (SLAM). To make it be feasibly achieved, previous works usually assume slow camera motions, which limits its usage in many real cases.
We propose an end-to-end 3D reconstruction system which combines color, depth and inertial measurements to achieve robust reconstruction with fast sensor motions.
Our framework extends Kalman filter to fuse the three kinds of information and involve an iterative method to jointly optimize feature correspondences, camera poses and scene geometry. 
We also propose a novel geometry-aware patch deformation technique to adapt the feature appearance in image domain, leading to a more accurate feature matching under fast camera motions.  
Experiments show that our patch deformation method improves the accuracy of feature tracking, and our 3D reconstruction outperforms the state-of-the-art solutions under fast camera motions.
\end{abstract}
%
%
\section{Introduction}
\label{sec:intro}
With the rapid development of capture and computation devices, such as depth sensors and GPUs, real-time 3D reconstruction has made big growth. In recent years, a lot of works have focused on indoor scene reconstruction. For example, InfiniTAM\cite{kahler2015very} only uses depth information to reconstruct 3D models, and estimate camera poses by an iterative closest point(ICP) algorithm\cite{rusinkiewicz2001efficient}. However depth only is extremely brittle in situations such as geometry-less scenes, bright windows and depth sensor noises, and can not eliminates accumulated errors. Drift-free camera tracking have been made breakthrough progress by monocular RGB-based methods, including direct methods\cite{forster2017svo} and feature point methods\cite{mur2015orb}. However these approaches can not reconstruct detailed and accurate 3D models. Further more, BundleFusion\cite{dai2017bundlefusion} and ElasticFusion\cite{whelan2016elasticfusion} use both color and depth information to estimate camera motions and generate 3D models based on implicit truncated signed distance fields(TSDFs) and surfel representation, respectively. 

Although these works exhibit reasonable results\cite{yan2014highly}, they still require strong assumptions, like static scene without dynamic object, sufficient texture and geometrical information, slow camera motions and invariant illumination. However these assumptions can not be satisfied in many applications.


	
In this paper, we make a step further by handling fast camera motions. For both color and depth, fast camera motion leads to large inter-frame distance, which makes it difficult to perform image feature matching (As images may be blurred and feature appearances may vary a lot.) and ICP based depth aligning. Similar with   Tristan et al.\cite{laidlow2017dense}, we solve the issue by introducing IMU information, gathered by an accelerometer and a gyroscope. Further combining with color and depth information, robust camera pose estimation and geometry fusion of an indoor scene are jointly achieved. The main contributions of our work are as follows:

(1) A \textit{RGB-D-inertial 3D reconstruction system based on extended Kalman filter framework}, which tightly combines the three kinds of information, and jointly achieves camera pose estimation and patch deformation in the kalman update step.

(2) A \textit{geometry-aware feature tracking method} for handling fast camera motion, which utilizes patch features to adapt blurry images and considers the deformation of patches in building feature matchings for images with very different perspectives.

(3) Through experiments on public datasets (including both synthetic and real data) and our data acquired by Intel Realsense ZR300, we see that our approach outperforms the state-of-the-art reconstruction systems under fast camera motions.
\begin{figure}[t]
\centering
\includegraphics[scale=0.2]{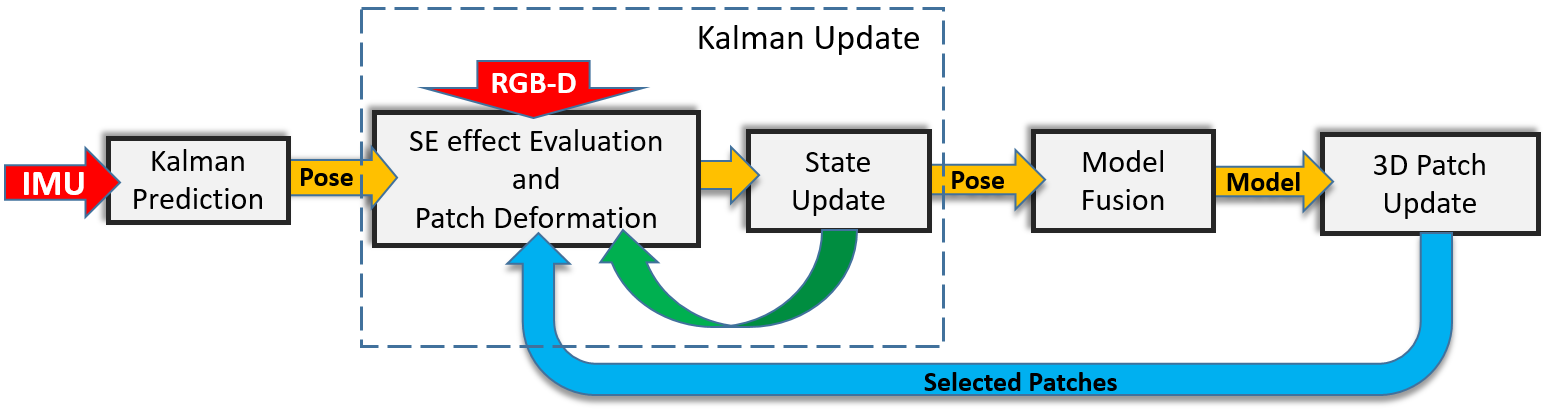}
\caption{Overview of our pipeline. The red, green and blue arrows represent the input acquired from current frame, iterative operation, and the patches from last frame.}
\label{fig:system}
\end{figure}

\section{Method}
The pipline of our system is illustrated in Fig. \ref{fig:system}. We introduce our method by following four parts, geometry-aware feature tracking which explores the SE effect and executes patch deformation, filter framework which explain the kalman prediction and update step, model fusion and patch update.

%
%
\subsection{Geometry-aware Feature Tracking}
Point-based feature tracking methods\cite{mur2015orb} will extract insufficient number of features when images are blurred or with less texture. Thus, patch-based method, which considers larger image regions, is proposed\cite{forster2017svo} to track features under these situations.
However, large patches may contain objects on different depth levels, which causes appearance changes in consecutive frames, especially when the camera motion is fast, leading to inaccurate feature tracking.
To address this problem, we combine color and depth information to back project 2D patches into 3D and re-project them to the camera of the next frame by using an initial camera motion. The projection helps us to deform the original patches to model the appearance changes, and the patch tracking can be easily and accurately achieved by the deformed patches.

\subsubsection{SE effect and Patch Deformation}
\label{sec:se}
\begin{figure}[!t]
\centering
\includegraphics[scale=.25]{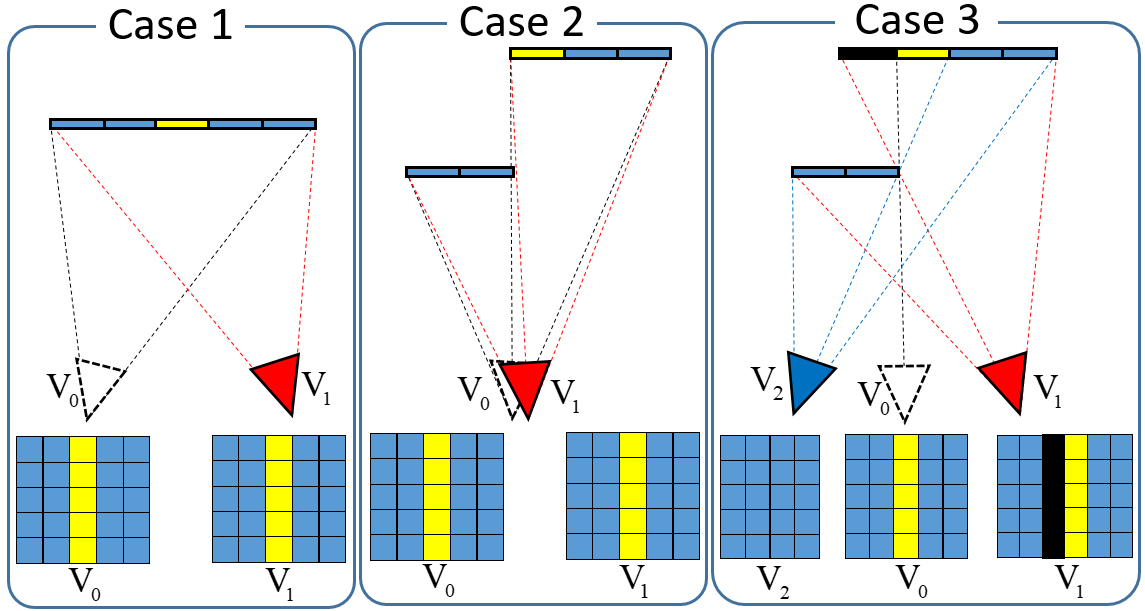}
\caption{This figure shows the patch SE effects caused by the camera motion and the geometrical shape of patches.}
\label{figure:patchdeform}
\end{figure}

When camera moves, a feature patch will be seen from different perspectives in different frames, and thus the shape and position of the feature patch in image coordinates vary from different frames. To account for the patch deformation, different from Bloesch et al.\cite{bloesch2017iterated} which only considers the 2D planar information of a patch, we use the 3D geometry of a patch to determine the 2D shape deformation between images recorded with fast camera motion.

Depending on different geometries of patches and irregularities of camera motions, patches may produce different deformations in consecutive frames. Fig. \ref{figure:patchdeform} shows three representative cases of patch deformations:
	
	\textbf{Case 1.} If there is no significant difference among pixel depths in a patch, then no matter how aggressive the camera motion is, the general shape of the patch will remain unchanged in two consecutive frames.
	
	\textbf{Case 2.} When camera moves slowly, the 2D shape of a patch will still remain unchanged even though there are large depth variances existing in the patch.
	
	\textbf{Case 3.} Different from case 2, if camera moves aggressively, the intensity distribution and shape of the patch will changes. If camera moves from $\textbf{V}_0$ to the viewpoint $\textbf{V}_2$, the yellow region will be occluded and the patch shrinks in the current frame. In addition, the black region which is occluded in $\textbf{V}_0$ is visible once the camera moves to $\textbf{V}_1$, and thus the patch shape extends in the current frame. These phenomenons are called shrink effect and extend effect, so we call them together by SE effect.

\begin{figure}[!tp]
\centering
\includegraphics[scale=.28]{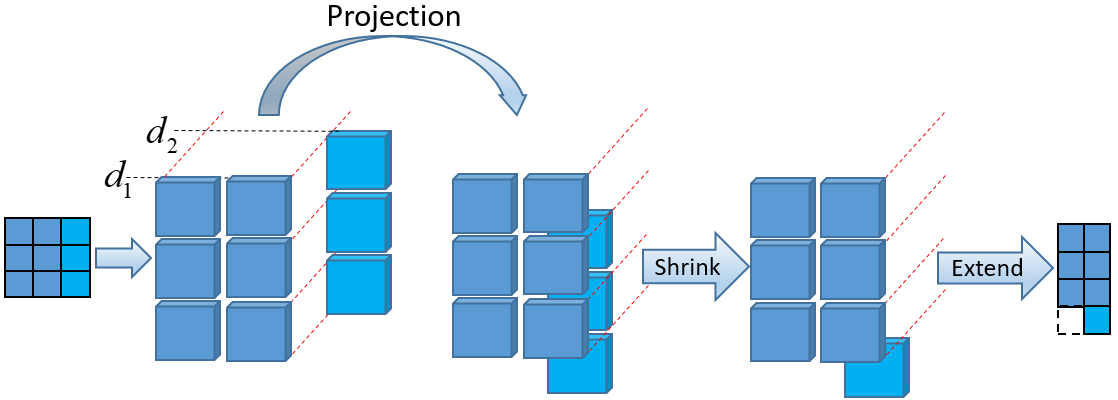}
\caption{Deformation process of SE patches.
}
\label{figure:patchwarping}
\end{figure}

We designed a unified deformation method to handle the SE effect. The details of patch deformation process are illustrated in Fig. \ref{figure:patchwarping} and formalized as follows:
	
Each pixel in a patch extracted from last frame $k-1$ are defined as $P^{k-1}_i := (\textbf{p}^{k-1}_i,I^{k-1}_i,d^{k-1}_i,\textbf{n}^{k-1}_i)$, where $\textbf{p}_i$ denotes the image coordinates of pixel $i$ in the patch. $I_i,d_i, \textbf{n}_i$ denote the intensity, depth and 3D normal of pixel $i$. $d_i$ and $\textbf{n}_i$ are obtained from the depth image, and since we have 3D information, we call our patches 3D patch features. We first back project each patch into 3D world coordinate system:
\begin{equation}
	\textbf{L}_i = \textbf{T}_{k-1}\pi^{-1}(P^{k-1}_i).
\end{equation}
Here, $\pi(\cdot)$ is to project a point from the 3D camera coordinate system to 2D pixel domain, and $\pi^{-1}(\cdot)$ represent the inverse operation. $\textbf{T}_{k-1}$ is the camera pose of frame $k-1$ which transforms a 3D point in the camera coordinate system of frame $k-1$ to the world coordinate system. Thus $\textbf{L}_i$ is in the world coordinate systme but is indexed from a pixel $i$ in frame $k-1$, so $\textbf{L}_i:=(\textbf{P}_i,I^{k-1}_i,\textbf{n}_i)$ where the intensity information does not change in the projection and $\textbf{P}_i$ indicates the 3D coordinate. Then we project it to the pixel coordinate system of the current frame $k$.
\begin{equation}
	P^{k'}_i = \pi(\textbf{T}^{-1}_k\textbf{L}_i).
\end{equation}
 Here $'$ means it is projected to this frame, not original from this frame. Note that $\textbf{T}_k$ is unknown and affects the 2D position of the projection. 
After projection, if two projected pixels happened to be in the same pixel coordinate, then we consider the shrink effect occurred in this patch. This usually happens when a region close to the camera covers the distant region. Therefore we remove the pixels corresponding to the distant region. 
Then we evaluate whether the extend effect happens or not. We set the shape of the projected patch to be the bounding box of all projected pixels. Then the extend effect will be verified if the height (or width) of the projected patch is greater than the original one. No matter which effect happens, we use $P^{k'}$ as the deformed patch in the following feature tracking steps.

\subsubsection{Objective}
In feature tracking, patches affected by SE effect are replaced with the corresponding projected patches. Then We track the projected patch features by both intensity and depth information.
	
Photometric error for each projected patch is computed as follows. We first extract the patch of the same size at the projected location of the current image as $P^{k}$, and then calculate the intensity difference between the extracted patch and the projected patch. The photometric error can be formalized as:
\begin{equation}
	E_p = \sum_i^Y \left \|
			I[P^{k}_i] - I[P^{k'}_i]
		 \right \|_2 ,
\end{equation}
where $Y$ denotes the number of pixels in the patch. $I$ indicates the corresponding intensity information of a patch. Then we compute point-to-plane geometry error as:
\begin{equation}
	E_g = \sum_i^Y \left \|
			\left (\textbf{P}[\textbf{T}_{k} \cdot \pi^{-1} (P^{k}_i )] 
			- \textbf{P}[\textbf{L}_i]
			\right ) \cdot \textbf{n}[\textbf{L}_i]
	\right \|_2.
\end{equation}
Given $E_p$ and $E_g$, the cost function for patch tracking is formulated as:
\begin{equation}
	E(\textbf{T}_k) = \lambda \sum_j^M E^j_p + (1-\lambda) \sum_j^M E^j_g ,
\label{equation:energy}
\end{equation}
where $j$ denotes the patch $j$, and $M$ indicates the number of patches.

\subsection{Filter Framework}
\label{sec:kalman}
Our EKF framework aims to tightly combine the color, depth and inertial measurements information. To be specific, we model the camera pose of each frame as the state of the EKF, and will solve it in the EKF. The observation of the EKF include both the color and depth images, and the relationship between the state and the observation is measured by the energy defined in Equation \ref{equation:energy}. If a state fits exactly to an observation, the energy is zero. On the other hand, the inertial information is used in the Kalman prediction step, which serves in building the motion prediction model. 

We follow the traditional Kalman filter to define the variables. A nonlinear discrete time system with state $\textbf{x}$, observation
term $z$, process noise ${\boldsymbol{\omega}}\sim N(0,{\textbf {Q}})$, and update noise ${\boldsymbol{\mu}}\sim N(0,{\textbf {U}})$ in $k$th frame can be written as
\begin{equation}
{\textbf {x}}_{k}=f({\textbf {x}}_{k-1},{\boldsymbol{\omega}}_{k}) ,
\end{equation}
\begin{equation}
{\textbf  {z}}_{{k}}=h({\textbf  {x}}_{{k}},{\boldsymbol{\mu}}_{{k}}) .
\end{equation}

In our framework, the state of the filter is composed of the following elements $\textbf{x} : \textbf{T}=(\textbf{R},\textbf{t})$, with a $3 \times 3$ camera rotation matrix $\textbf{R}$ and a $3 \times 1$ camera translation vector $\textbf{t}$ related to the world coordinate system which is assigned to be the camera coordinate system of the first frame.In the following, the superscript symbol '$+$' denotes a-posteriori estimate of a variable calculated from the Kalman update step and '$-$' denotes a-prior estimate from the Kalman prediction step.

\subsubsection{Kalman Prediction and State Propagation}
	Given an a-posteriori estimate $\textbf{x}^+_{k-1}$ with covariance $\textbf{P}^+_{k-1}$, the prediction step of the EKF yields a-priori estimate at the next frame: 
\begin{equation}
{\textbf {x}}^-_{k}=f(\textbf {x}^+_{k-1},0) ,
\end{equation}	
\begin{equation}
{\textbf {P}}^-_{k}=
{\textbf {F}}_{k}{\textbf {P}}^+_{k-1}{\textbf {F}}_{k}^{T}+{\textbf {Q}}_{k} ,
\end{equation}	
with the Jacobians:
\begin{equation}
{\textbf {F}}_{k}=
\left.{\frac {\partial f}{\partial {\textbf {x}}}}\right\vert _{\textbf {x}^+_{k-1},0} .
\end{equation}

The key in the Kalman prediction step is to define the function $f$. In our EKF framework, the inertial measurements are employed in the definition.
Following \cite{mourikis2007multi}, we get the actual sensor acceleration $\textbf{a}$ and angular velocity $\textbf{w}$ from inertial measurements. We assume that the IMU is synchronized with the camera and acquires measurements with time interval $\tau $ which is much smaller than that of the camera. Hence, we denote $N$ as the number of inertial measurements acquired in two consecutive camera frames, and then merge them together by pre-integration method\cite{forster2017manifold} to predict camera rotation $\Delta \textbf{R}$ and translation $\Delta \textbf{t}$ between two consecutive camera frames:
\begin{equation}
	\Delta \textbf{R} = \Phi \cdot \prod_{n=1}^N \mathrm{Exp}(\textbf{w}_n \cdot \tau) 
\end{equation}
\begin{equation}
	\Delta \textbf{v} = \sum_{n=1}^N \Delta \textbf{R}_n \cdot \textbf{a}_n \cdot \tau 
\end{equation}
\begin{equation}
	\Delta \textbf{t} = \Phi \cdot \sum_{n=1}^N (\Delta \textbf{v}_n \cdot \tau + \frac{1}{2} \textbf{g} \cdot \tau^2 + \frac{1}{2}  \Delta \textbf{R}_n \cdot \textbf{a}_n \cdot \tau^2) .
\end{equation}	
In the above three equations, the subscript '$n$' denotes the corresponding variable at the $n$th IMU input in consecutive camera frames. Besides, $\Delta \textbf{v}$ is the accumulated IMU linear velocity from the last camera frame to the current camera frame, and $\Phi$ is the extrinsic matrix from the IMU coordinate to the camera coordinate. $\textbf{g}$ is the gravity acceleration, and $\mathrm{Exp}(\cdot)$ denotes the exponential map from Lie-algebra to Lie-group. 
Details about this predition step can be found in  \cite{mourikis2007multi}.
 Finally, the states predicted in current frame $k$ can be formulated as:
\begin{equation}
\begin{bmatrix}
\textbf{R}_k^- & \textbf{t}_k^-\\ 
\textbf{0} & 1
\end{bmatrix} = 
\begin{bmatrix}
\Delta \textbf{R}_k \cdot \textbf{R}_{k-1}^+ & \Delta \textbf{t}_k + \textbf{t}_{k-1}^+\\ 
\textbf{0} & 1
\end{bmatrix} .
\end{equation}

\subsubsection{Kalman Update and Iteration}
In traditional extended Kalman update step, the measurement residual is modeled as:
\begin{equation}
	\textbf{y}_k = \textbf{z}_k - h({\textbf  {x}}_{k}^-,{\textbf{0}}).
\end{equation}
Here, $\textbf{0}$ means we directly use ${\textbf  {x}}_{k}^-$ to calculate the residual without adding any Gausian noise. The updated state is formulated as:
\begin{equation}
	\textbf{x}_k^+ = \textbf{x}_k^- + \textbf{K}_k \cdot \textbf{y}_k ,
	\label{equation:x_k+}
\end{equation}
where $\textbf{K}_k$ is the Kalman gain. In our method, we defined the residual as the photometric and geometric error of patches (equation \ref{equation:energy}), thus the residual can be formulated as:
\begin{equation}
\textbf{y}_k = 0- E(\textbf{T}^-_k)=-E \left(
\begin{bmatrix}
\textbf{R}_k^- & \textbf{t}_k^-\\ 
\textbf{0} & 1
\end{bmatrix} \right) .
\label{equation:y_k}
\end{equation}
Notice that the deformations of the SE patches, which are used in calculating $\textbf{y}_k$ by Equation \ref{equation:y_k}, is heavily affected by the camera poses. So after we obtained an updated camera pose by Equation \ref{equation:x_k+}, we use the newly updated camera pose to iteratively calculate the deformations of the SE patches and refine the camera poses by Equation \ref{equation:x_k+} again. In this manner, we can estimate a more accurate $\textbf{x}_k^+$. To be more specific, we use $m$ to denote the iterations, and thus we have:
\begin{equation}
h(\textbf{x}_{k,m}^+,0) 
=
E \left(
\begin{bmatrix}
\textbf{R}_{k,m}^+ & \textbf{t}_{k,m}^+\\ 
\textbf{0} & 1
\end{bmatrix} \right) ,
\end{equation}
and the Kalman gain respect to each iteration is:
\begin{equation}
\textbf{K}_{k,m} = \textbf{P}^-_{k} \textbf{H}^T_{k,m} \textbf{S}^{-1}_{k,m}
\end{equation}
\begin{equation}
\textbf{S}_{k,m} = \textbf{H}_{k,m} \textbf{P}^-_{k} \textbf{H}^T_{k,m} + \textbf{U}_k.
\end{equation}
As defined in the begin of section \ref{sec:kalman}, $\textbf{U}_k$ is the covariance matrix of noise $\boldsymbol{\mu}_k$. And the Jacobians updated in every iteration are formulated as:
\begin{equation}
{\textbf {H}}_{k,m}=
\left.{\frac {\partial h}{\partial {\textbf {x}}}}\right\vert _{\textbf {x}^+_{k,m},0}
\end{equation}
Then the updated state of each iteration is calculated as follows:
\begin{equation}
\textbf{x}_{k,m+1}^+ = \textbf{x}_{k,m}^+ - \textbf{K}_{k,m} \cdot h({\textbf {x}}_{k,m}^+,{\textbf{0}}).
\end{equation}	
Notice that $\textbf{x}_{k,0}^+$ is set to be $\textbf{x}_k^-$. 
Finally, the iteration is terminated when the absolute value of $\textbf{K}_{k,m} \cdot h({\textbf {T}}_{k,m}^+,{\textbf{0}})$ is below a certain threshold and the covariance matrix is only updated once the process has converged after $\eta$ iterations:
\begin{equation}
\textbf{P}^+_k = (\textbf{I}-\textbf{K}_{k,\eta} \textbf{H}_{k,\eta}) \textbf{P}^-_k
\end{equation}

\subsection{Model Fusion and Patch Update}
We use the volumetric truncated signed distance function (TSDF)\cite{curless1996volumetric} to incrementally fuse each consecutive depth frame $\textbf{D}_k$ into one 3D geometry model $\textbf{M}_k(\textbf{X})$, with the associated camera pose from Kalman update $\textbf{R}^+_k,\textbf{t}_k^+$.
Details about depth fusion can be found in\cite{zhang2017mixedfusion}.

After the reconstruction, we should update patch features for subsequent tracking. We get rid of bad features based on average pixel intensity error, and re-extract squared patch feature for those with non-square shapes affected by the SE effect. Then, we add new features with distinct intensity gradient and sufficient depth information evaluated by FAST corner detector\cite{rosten2006machine} and the number of pixels with available depth information. Finally, patch intensity information is updated by current color and depth information is acquired from the 3D geometry model which has better quality than the current depth image.

\section{Experiments}
We first demonstrate the effectiveness of our geometry-aware feature tracking method, which evaluates SE effect and deforms patches for accurate feature tracking in sequences with fast camera motion. 
Then, we evaluate the benefits of inertial information by comparing our system with and without IMU.
Finally, our 3D reconstruction method is compared against state-of-the-art systems in datasets with fast sensor motion.
\begin{figure}[!tb]
\centering
\includegraphics[scale=0.2]{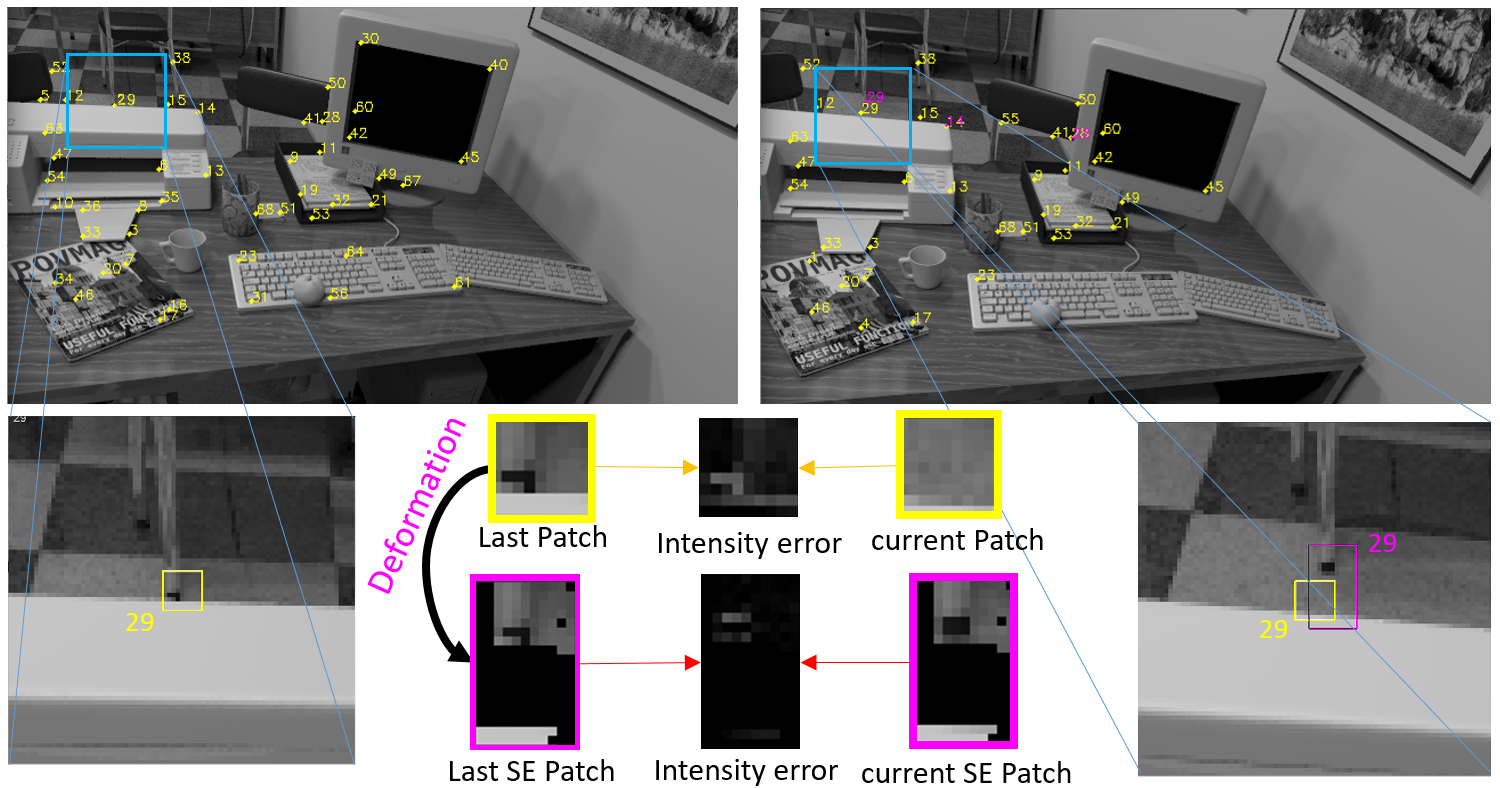}
\caption{This figure shows the feature tracking results of ours and direct method. }
\label{fig:patchsecompare1}
\end{figure}

\renewcommand\arraystretch{1.5}
\begin{table}
\caption{Comparison of Patch Feature Tracking}
\centering
\begin{tabular}{|c|c|c|c|}  
\hline                      
\multirow{2}{*}{Type} &
\multirow{2}{*}{Dataset} &
\multicolumn{2}{c|}{AIE}\\
\cline{3-4}
&&DM &Ours\\
\hline
\multirow{3}{*}{slow}&
 TUM\_freiburg1\_desk & 13.3756 & \textbf{9.53}\\
 & ICL\_NUIM\_lr\_kt2 & 4.8981 & 4.0825\\
 & Dorm\_slow & 8.1312 & 7.8934\\
\hline
\multirow{2}{*}{fast}&
 ICL Fast Motion & 17.219 & \textbf{7.8328}\\
 & Dorm\_fast & 13.9011 & \textbf{7.9325}\\
\hline
\end{tabular}
\label{table:compare}
\end{table}

\subsection{Evaluation}
\textbf{Feature tracking.} We compare our feature tracking method against traditional direct method which does not take the SE effect under consideration.
In order to achieve equitable comparison, we use a patch-size $10 \times 10$ in both methods and extract no more than 100 patches in each frame. 
Fig. \ref{fig:patchsecompare1} shows the tracking results of a patch feature in two consecutive frames. The tracking result of traditional method is severely influenced by SE effect and got bad intensity error, while our method deforms the patch and eliminates the influence caused by SE effect to get lower intensity error.

We compare on several datasets, contains ICL datasets\cite{handa2012real,handa:etal:ICRA2014}, TUM datasets\cite{sturm12iros} and our datasets gathered by a handheld sensor. The average intensity error(AIE) of patches are listed in Table \ref{table:compare}.
All datasets are divided into slow and fast depending on the qualities of recorded images. To be more specific, as there is no explicit criteria for dividing camera speed, thus we empirically set, based on the unified characteristics of most public datasets, the motion without creating image motion blur as slow camera motion, and the motion which creates severe image blur as fast camera motion. From the table, we find that our method gets lower AIE in all datasets, especially in datasets with fast camera motion.

\begin{figure}[!tb]
\centering
\includegraphics[scale=0.24]{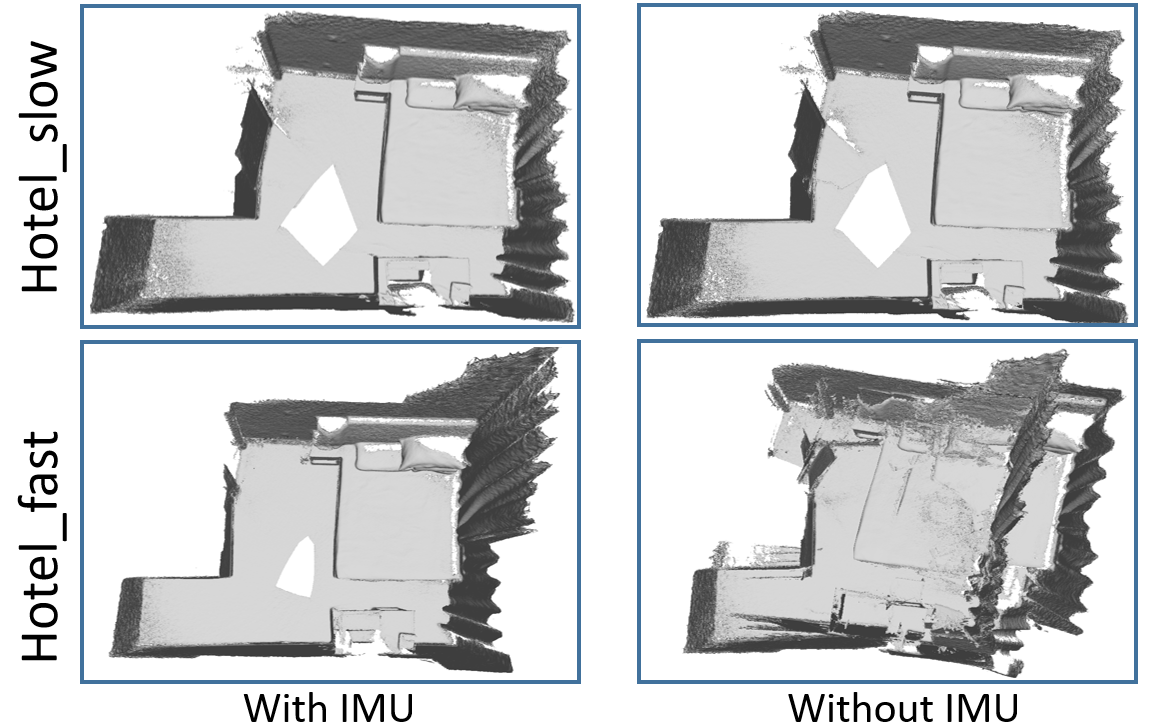}
\caption{The reconstruction results of a hotel under slow and fast camera motion.}
\label{fig:imucompare}
\end{figure}

\begin{figure}[!tb]
\centering
\includegraphics[scale=0.26]{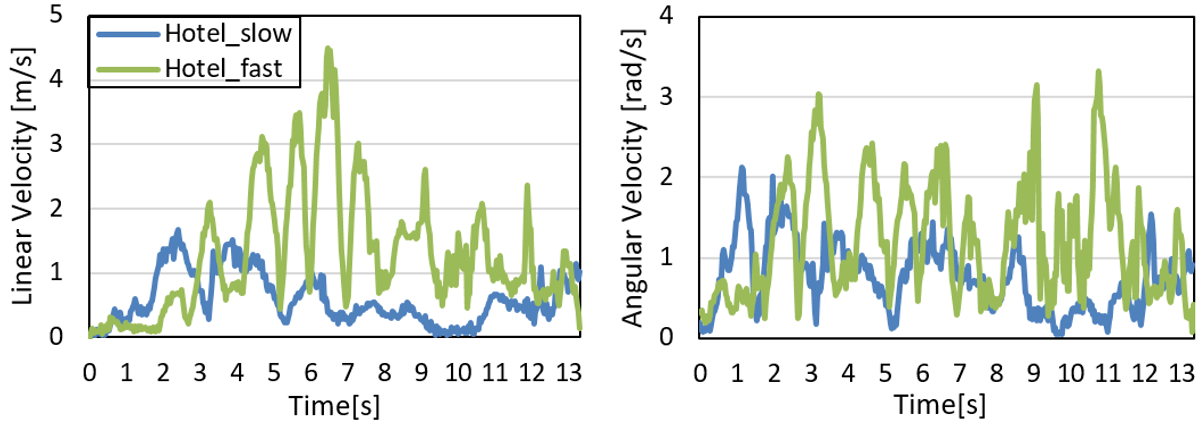}
\caption{Camera linear and angular velocity of the two sequences used in IMU evaluation.}
\label{fig:speed}
\end{figure}

\textbf{IMU evaluation.} To verify whether the integration of IMU helps to reconstruct the scene geometry during fast camera motion, we compare the results with and without IMU on two datasets with slow and fast camera motions, respectively. As shown in Fig. \ref{fig:imucompare}, on the dataset with slow camera motion, the system without IMU works on-par with the complete system, while it fails to reconstruct the model for fast camera motions.
Fig. \ref{fig:speed} demonstrates the details of camera motions in the two datasets. From the figure, we find that in the fast dataset, there exist some subsequences with large linear and angular velocities of camera, which cause the system without IMU fails to track camera poses. Notice that other fast datasets used in our experiments also contain this kinds of subsequences. 

\begin{figure}[!tb]
\centering
\includegraphics[scale=0.35]{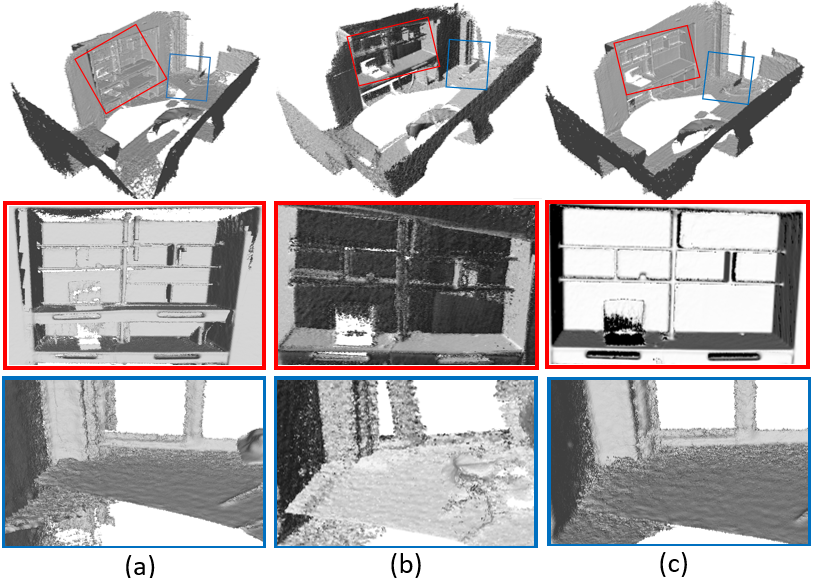}
\caption{Comparison of reconstruction with fast camera motion in (a)InfiniTAM (b)ElasticFusion and (c)Ours.}
\label{fig:fast recons compare}
\end{figure}
\subsection{Comparison}
We compare our 3D reconstruction systems with InfiniTAM\cite{kahler2015very}, a typical voxel based scene reconstruction method, Bundlefusion\cite{dai2017bundlefusion} which proposed an efficient global pose optimization algorithm, and a surfel based method ElasticFusion\cite{whelan2016elasticfusion} which contains loop closure and executes model refinement through non-rigid surface deformations.

The results of sequence Dorm\_fast which reconstruct the entire scene are exhibited in Fig. \ref{fig:fast recons compare}. As BundleFusion fails once the camera speeds up and subsequently restarts when the camera slows down, thus we only show its reconstruction precess in our supplementary video https://www.youtube.com/watch?v=Jy3SGqWuhp8. From the Fig. \ref{fig:fast recons compare} we find that InfiniTAM can not maintain consistency of the reconstructed geometry, which is mainly caused by the inaccurate camera pose estimation and the large accumulated error. Meanwhile, the loop closure function of ElasticFusion, aiming to eliminate accumulated error, is always invalid in fast camera motion, and thus leads to the fail reconstruction of the parts shown with red and blue bounding boxes. In the opposite, our system reconstructs a good geometry of the scene even without loop closure.

We encourage the reader to watch our video for a better visualization of comparison results.

\section{Conclusion and Future Work}
We present a real-time system for indoor scene reconstruction by tightly-coupled RGB-D-Inertial information with an extended Kalman filter. The key feature of our method is that it can estimate camera pose and reconstruct 3D scene model with fast camera motion. In addition, we explore the SE effect and propose a geometry-aware patch deformation method to eliminate the influence during feature tracking.
However, our system has not achieved loop closure with fast camera motion. The reason is that the degraded image information caused by fast camera motion, such as image motion blur, results in the difficulties in loop detection(or feature association) of loop closure method. In future work, we wish to address the problem of loop closure under fast camera motion.


\bibliographystyle{IEEEbib}
\bibliography{FastFusion}

\end{document}